\documentclass[final,3p]{CSP}
\usepackage{amssymb}
\usepackage{amsmath}
\usepackage{changepage}
\usepackage{graphicx}
\usepackage{caption}
\usepackage{booktabs}
\usepackage{threeparttable}

\usepackage{xcolor}

\begin{document}

\begin{frontmatter}

\title{Using Trust in Automation to Enhance Driver-(Semi)Autonomous Vehicle Interaction and Improve Team Performance}

\author[mymainaddress]{\underline{Hebert Azevedo-Sa}}
\author[mymainaddress]{Xi Jessie Yang}
\author[mymainaddress]{Lionel P. Robert Jr.}
\author[mymainaddress]{Dawn M. Tilbury}

\address[mymainaddress]{University of Michigan-Ann Arbor}


\end{frontmatter}

\section{Introduction}
\noindent
Trust in robots has been gathering attention from multiple directions \cite{Chen2018, Korber2018, Sheridan2019, Kundinger2019, robinette2019exploring, lee2015trust, xie2019robot}, as it has a special relevance in the theoretical descriptions of human-robot interactions \cite{steinfeld2006common, billings2012human}.
It is essential for reaching high acceptance and usage rates of robotic technologies in society \cite{Metcalfe2017}, as well as for enabling effective human-robot teaming \cite{robinette2019exploring, azevedo2020comparing, azevedo2020context, azevedo2020real, azevedo2021handling, azevedo2021internal}.
Researchers have been trying to model the development of trust in robots to improve the overall ``rapport'' between humans and robots.
In particular, researchers sought to characterize factors that induce changes in drivers' trust in automated driving systems (ADSs)---the systems that provide the autonomous capabilities of (semi)autonomous vehicles---over their interactions \cite{Zhao2019}.

Trust in automation \cite{Muir1987, Lee1994} is a popular concept among researchers, and evolved from the theory behind interpersonal trust development \cite{barber1983logic, Rempel1985}.
Miscalibration of trust in automation is a common issue that jeopardizes the effectiveness of automation use \cite{christensen2019reducing}.
It happens when a user's trust levels are not appropriateate to the capabilities of the automation being used.
Users can be: \textit{undertrusting} the automation---when they do not use the functionalities that the machine can perform correctly because of a ``lack of trust''; or \textit{overtrusting} the automation---when, due to an ``excess of trust'', they use the machine in situations where its capabilities are not adequate \cite{MUIR1996, Hoff2013, Hu2018}.
The influence of different risk types (internal, from the ADS itself, and external, from the environment) \cite{petersen2018influence} and of situational awareness in trust development has been investigated \cite{petersen2017effects}. 
Different risk types were found to influence trust development and to moderate the impacts of trust on non-driving-related task (NDRT) performance and on trusting behaviors. Moreover, when the ADS provides high quality information, drivers are able to increase their trust in the automation and achieve better performance on NDRTs.

The main objective of this work is to examine driver's trust development in the ADS.
We aim to model how risk factors (e.g.: false alarms and misses from the ADS) and the short term interactions associated with these risk factors influence the dynamics of drivers' trust in the ADS.
The driving context facilitates the instrumentation to measure trusting behaviors, such as drivers' eye movements and usage time of the automated features.
Our findings indicate that a reliable characterization of drivers' trusting behaviors and a consequent estimation of trust levels is possible. 
We expect that these techniques will permit the design of ADSs able to adapt their behaviors to attempt to adjust driver's trust levels.
This capability could avoid under- and overtrusting, which could harm their safety or their performance.

\section{Methodology}
\noindent
We seek to characterize trust dynamics in short term interactions, i.e., during the different events that occur when drivers operate vehicles featuring ADSs.
In other words, we aim to describe how trust in the ADS changes immediately after the ADS meets (or fails to meet) drivers' expectations.
For this we have quantified drivers' trust in the ADS at specific time instances and used linear mixed effects models \cite{swaminathan2008estimation} to establish a state-space model for trust dynamics (\ref{eq:ss_trust_model}),
\begin{equation}
\label{eq:ss_trust_model}
\begin{aligned}
  T(t+1) = \mathbf{A} T(t) + \mathbf{B}
  \begin{bmatrix}
    L(t) \\
    M(t) \\
    F(t) 
  \end{bmatrix}
  + u(t) \,; \,\,\,\,\,
  \begin{bmatrix}
    \varphi(t) \\ 
    \pi(t) \\
    \upsilon(t) 
  \end{bmatrix} = \mathbf{C} T(t) + w(t) \,,
\end{aligned}
\end{equation}
where $T$ represents drivers' self-reported trust; 
$\mathbf{A}$, $\mathbf{B}$ and $\mathbf{C}$ represent the state-space model matrices; 
The time variable $t$ indexes an interaction event, characterized by the occurrence of true alarms (when the ADS is able to identify an obstacle on the road), false alarms (when the ADS issues an alarm but there is no obstacle on the road) and misses (when the ADS is unable to identify an obstacle on the road).
These events are respectively represented by $L(t)$, $F(t)$ and $M(t)$, which are inputs of the model. 
$\varphi$, $\pi$ and $\upsilon$ represent, respectively, the measures of visual focus (measured from eye-tracking), NDRT performance and automation usage (observations);
$u(t)$ and $w(t)$ represent random zero-mean noises intended to capture drivers' individual characteristics and biases.
With this model, we develop a trust estimator able to provide real-time estimates of trust $T(t)$.

\begin{figure}[!ht]
    \centering
    \includegraphics[width=0.8\linewidth]{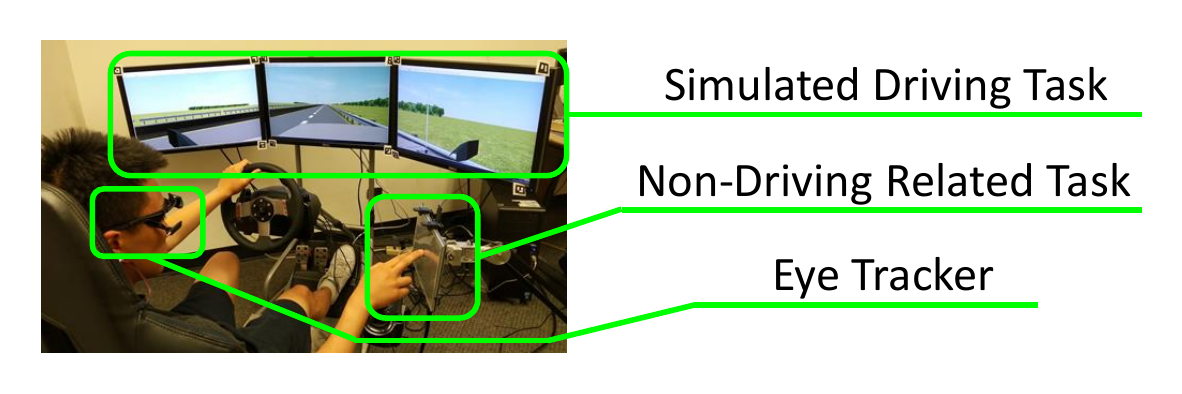}
    \caption{The setup for user experiments. Drivers interact with a simulated autonomous vehicle while performing a NDRT and reporting their trust in the system.}
    \label{fig:exp_setup}
\end{figure}

For data acquisition, we use an autonomous vehicle simulator in a user experiment where we expose the participants to different behaviors from the ADS.
The participants experience two trials, i.e., in straight and curvy roads. 
Within each of the two trials there are $12$ interaction events.
We request the participants to self-report their trust in the ADS and measure the real time variables that represent their trusting behaviors.
We use Muir's trust questionnaire \cite{MUIR1996} and scale the self-reported scores to match the interval from $1$ to $100$ points. Figure \ref{fig:exp_setup} shows our experimental setup.

\section{Results}
\begin{figure}[!ht]
  \centering
  \includegraphics[width=1\linewidth]{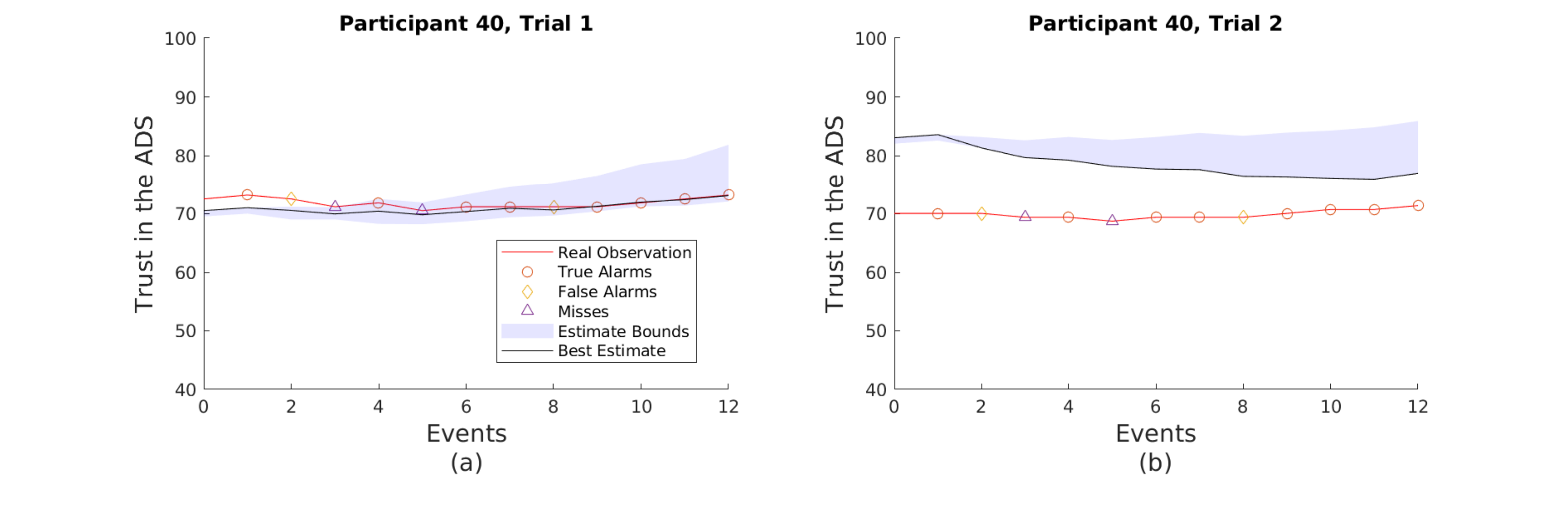}
  \caption{Curves representing the self-reported trust levels and the calculated estimates of trust for the $40^{th}$ participant of the experiment, over 12 events. The events could be true alarms, false alarms, or misses. The blue shaded bands indicate the approximate bounds for the estimates.}
  \label{fig:results}
\end{figure}

\noindent
We used the data acquired from $80$ participants to compute the parameters for the trust dynamics model (\ref{eq:ss_trust_model}). 
Table \ref{tab:ss_parameters} presents the state-space model matrices.
With the identified parameters we designed the trust estimator, which is based on a linear Kalman filter.
Figure \ref{fig:results} presents the trust estimation results for a randomly chosen participant, bringing together their two trials and the different estimation results for the $12$ events within each trial.

The trust estimate bounds represented in Figure \ref{fig:results} are approximations obtained with the juxtaposition of $100$ simulated estimates. 
The estimates vary with the noise introduced in the Kalman filter.
This variability is due to the uncertainty represented by the random noise parameters $u$ and $w$.
The best estimate and the real observation curves are also represented in the plots.

\begin{table}
\renewcommand{\arraystretch}{1.3}
\caption{State-space model parameters for trust in the ADS dynamics}
\label{tab:ss_parameters}
\centering
\begin{tabular}{c|c|c}
    \hline
    Parameter  &  Value Estimate & S.E.M\\
    \hline
    \hline

    $\mathbf{A}$    &   $[1.00]$ & $[0.25]$\\
    \hline
    
    $\mathbf{B}$    &   $\begin{bmatrix} 
                                0.224 & -0.670 & -0.798 \\
                           \end{bmatrix}$ 
                           & 
                           $\begin{bmatrix} 
                                0.079 & 0.084 & 0.083 \\
                           \end{bmatrix}$ 
                           \\
    \hline

    $\mathbf{C}$    &   $\begin{bmatrix} 
                                7.01 \times 10^{-3}  \\
                                4.23 \times 10^{-3}  \\
                                9.20 \times 10^{-3} \\
                           \end{bmatrix}$ 
                           & 
                           $\begin{bmatrix} 
                                3.6 \times 10^{-4}  \\
                                1.3 \times 10^{-4}      \\
                                1.0 \times 10^{-4} \\
                           \end{bmatrix}$
                           
                           \\
    \hline

    $\sigma^2_{u}$  &   $0.26$ & --\\
    \hline
    
    $\mathbf{\Sigma}_{w} = \begin{bmatrix} 
                                \sigma^2_{\varphi} & 0 & 0  \\
                                0 & \sigma^2_{\pi} & 0      \\
                                0 & 0 & \sigma^2_{\upsilon} \\
                           \end{bmatrix}$
                    &   $\begin{bmatrix} 
                                0.18 & 0 & 0  \\
                                0 & 0.07 & 0      \\
                                0 & 0 & 0.06 \\
                           \end{bmatrix}$ 
                    & --\\
    \hline
    \hline

\end{tabular}
    \begin{tablenotes}
      \small
      \item \hspace{1.8cm} Note: S.E.M = Standard Error of the Mean.
    \end{tablenotes}
\end{table}

Our results show that the proposed method can successfully track the self-reported trust levels from the observation variables ($\varphi$, $\pi$ and $\upsilon$). 
However, comparing the curves from Figures \ref{fig:results}(a) and \ref{fig:results}(b), it is notable that the success of the estimation process highly depends on a precise initial estimate of trust in the ADS. 
In Figure \ref{fig:results}(b), the difference between curves got reduced over the interactions, but the total of $12$ events was not sufficient for the trust estimator to track the observed self-reported trust levels.
Two causes contribute for this: first, the variation of drivers' trust in the ADS is a slow process, which can be noted by the flatness of the red curves in Figure \ref{fig:results}; second, the noise terms ($u$ and $w$) have large variances, as they intentionally capture the individual differences of behaviors from each participant. 
These large variances slow down the convergence of the curves, as the Kalman filter structure has a conservative behavior when the observations are contaminated by noise.
This limitation might be addressed by establishing a longer term and individualized training process for the trust estimator, capable of providing more precise initial trust estimates. 
Furthermore, this process might also be used for tuning the parameters of the model, which would improve the accuracy of the estimates.

\section{Conclusion}
\noindent
In this work we explored how trust develops over interactions between drivers and ADSs, and how the reliability of the ADS affects this trust development.
We quantified trust and introduced an estimation method to describe and track trust dynamically in the short term interactions represented by true alarms, false alarms and misses.
This method might be used for the design of ADSs that could adapt their behaviors to avoid trust miscalibration and improve team effectiveness, focusing on increasing safety and performance in the driving context.




\bibliographystyle{bibft}\it
\bibliography{bibfile}



\end{document}